\documentclass{article}

\usepackage[final]{neurips_2022}

\usepackage[utf8]{inputenc} 
\usepackage[T1]{fontenc}    
\usepackage{hyperref}       
\usepackage{url}            
\usepackage{booktabs}       
\usepackage{amsfonts}       
\usepackage{nicefrac}       
\usepackage{microtype}      
\usepackage{xcolor}         

\usepackage{amsmath}
\usepackage{tikz}
\usepackage{pgfplots}
\usepackage{amssymb}
\usepackage{amsthm}

\title{A Simple But Powerful Graph Encoder for Temporal Knowledge Graph Completion}

\author{Zifeng Ding$^{1,2}$, Yunpu Ma$^{1,2}$, Bailan He$^{1}$, Zhen Han\thanks{\; Corresponding author.}$\;^{1,2}$, Volker Tresp$^{*1,2}$ \\
$^{1}$Institute of Informatics, LMU Munich $\;$  $^{2}$ Corporate Technology, Siemens AG\\
\texttt{\{zifeng.ding, bailan.he, zhen.han\}@campus.lmu.de}\\
\texttt{cognitive.yunpu@gmail.com, volker.tresp@siemens.com}\\}

\begin{document}

\maketitle

\begin{abstract}
Knowledge graphs contain rich knowledge about various entities and 
    the relational information among them, while temporal 
    knowledge graphs (TKGs) describe and model the interactions of the entities over 
    time. In this context, automatic temporal knowledge graph 
    completion (TKGC) has gained great interest. Recent TKGC methods integrate advanced deep learning techniques, e.g., Transformers, and achieve superior model performance. However, this also introduces a large number of excessive parameters, which brings a heavier burden for parameter optimization. In this paper, 
    we propose a simple but powerful 
    graph encoder for TKGC, called TARGCN. TARGCN is parameter-efficient, and it extensively explores every entity's temporal context for learning contextualized representations. We find that instead of 
    adopting various kinds of complex modules, it is more beneficial to efficiently capture the temporal contexts of entities. We experiment TARGCN on three 
    benchmark datasets. Our model can achieve a more than 46\% relative improvement 
    on the GDELT dataset compared with state-of-the-art TKGC models. Meanwhile, it outperforms the strongest baseline on the ICEWS05-15 dataset with around 18\% fewer parameters.
\end{abstract}

\section{Introduction}
\label{Introduction}
A \textit{Knowledge Graph} (KG) is a graph-structured \textit{Knowledge Base} (KB) that stores relational facts. KGs have drawn increasing research interest since they serve as key drivers for a wide range of downstream tasks in artificial intelligence, e.g., question answering \cite{DBLP:journals/corr/abs-2208-06501}, commonsense reasoning \cite{DBLP:conf/acl/XingSMLMW20}, and recommender systems \cite{DBLP:conf/aaai/WangWX00C19}. A fact in a KG is described as a triplet $(s,r,o)$, e.g., (\textit{Joe Biden}, \textit{is president of}, \textit{USA}), where $s$, $o$, $r$ denote the subject entity, the object entity, and the relation between $s$ and $o$. While KGs contain rich knowledge about entities and the relational information among them, they do not consider the nature of ever-evolving relational facts over time. For example, consider a KG triplet (\textit{Donald Trump}, \textit{is president of}, \textit{USA}). According to world knowledge, this triplet is valid only before \textit{Joe Biden} took the place of \textit{Donald Trump} as the president of the \textit{USA}. This implies a shortcoming of KGs and calls for the introduction of \textit{Temporal Knowledge Graphs} (TKGs). In TKGs, every fact is augmented with a specific timestamp $t$ such that it can be described with a quadruple $(s,r,o,t)$. In this way, every fact in TKGs has its own time validity and this enables TKGs to capture the factual information in a time-varying context.

\textit{Temporal Knowledge Graph Completion} (TKGC) is a task aiming to infer the missing facts in TKGs. There exist two lines of TKGC methods. ($1$) A lot of prior methods attempt to incorporate temporal information into the existing KG reasoning scoring models and build novel time-aware score functions for TKGs \cite{DBLP:conf/www/LeblayC18,DBLP:conf/emnlp/Garcia-DuranDN18,Ma2019EmbeddingMF,DBLP:conf/iclr/LacroixOU20,DBLP:journals/corr/abs-2109-08970}.
($2$) Another line of work takes advantage of neural structures, e.g., \textit{Graph Neural Networks} (GNNs) \cite{DBLP:conf/icml/NiepertAK16,DBLP:conf/iclr/KipfW17} and recurrent models, for modeling the temporal information in TKGC \cite{DBLP:conf/emnlp/WuCCH20,DBLP:conf/kdd/JungJK21}. Experimental results show that neural structures help to achieve state-of-the-art performance on the TKGC task. However, employing additional neural structures on top of the existing KG score functions normally leads to a higher number of model parameters. The parameter consumption increases even more when these models are equipped with advanced deep learning modules, e.g., attention mechanisms and Transformers \cite{DBLP:conf/nips/VaswaniSPUJGKP17}, thus causing high memory consumption and bringing a heavier burden for parameter optimization.

In this paper, we follow the trend of the second line of methods, aiming to design a neural-based graph encoder for TKGC that helps to cut the parameter consumption and the model complexity while maintaining superior model performance. We propose a time-aware relational graph encoder: \textit{\textbf{T}ime-\textbf{a}ware \textbf{R}elational \textbf{G}raph \textbf{C}onvolutional \textbf{N}etwork} (TARGCN). We find that our light-weighted time-aware relational graph encoder performs well on the TKGC task, and it requires relatively few parameters. The contribution of our work can be summarized as follows:
(i) We propose a time-aware relational graph encoder, i.e., TARGCN, for the TKGC task. TARGCN learns an entity's time-aware representation by sampling a temporal neighboring graph which consists of extensive temporal neighbors, and encodes temporal information by modeling time differences with a functional time encoder.
(ii) To test the robustness of TKGC models on irregular timestamped data, we propose a new dataset ICEWS14-irregular. TARGCN achieves superior performance on it compared with several recently proposed TKGC methods. Besides, TARGCN outperforms previous methods with a huge margin in predicting the links at unseen timestamps, which also shows its strong robustness.
(iii) TARGCN serves as a parameter-efficient model. To achieve the same performance, it requires much fewer parameters compared with two recently proposed neural-based TKG reasoning models, TeMP \cite{DBLP:conf/emnlp/WuCCH20} and T-GAP \cite{DBLP:conf/kdd/JungJK21}. 
(iv) We evaluate TARGCN on three benchmark TKGC datasets. It achieves superior performance on all datasets. On the GDELT \cite{leetaru2013gdelt} dataset, it achieves a more than 46\% relative improvement compared with the best baseline.

\section{Preliminaries and related work}
\paragraph{Knowledge graph embedding models.} \textit{Knowledge graph embedding} (KGE) models have shown great success in KG reasoning tasks. TransE \cite{DBLP:conf/nips/BordesUGWY13} is the first KGE model that introduces translational embeddings into KG representation learning. Many further works \cite{DBLP:conf/aaai/LinLSLZ15,DBLP:conf/iclr/SunDNT19,DBLP:conf/nips/AbboudCLS20} are inspired and extend the relational translations in different spaces to capture complex relational information. Another line of KGE methods are tensor factorization-based models \cite{DBLP:conf/icml/NickelTK11,DBLP:journals/corr/YangYHGD14a,DBLP:conf/emnlp/BalazevicAH19}. They encode entity and relation embeddings as vectors and then use bilinear functions to compute the plausibility scores for KG facts. Besides, neural-based relational graph encoders have been rapidly developed and have shown great power in capturing structural information of KGs. R-GCN \cite{DBLP:conf/esws/SchlichtkrullKB18} incorporates relation information into a \textit{Graph Convolutional Network} (GCN) \cite{DBLP:conf/iclr/KipfW17} to enable relational reasoning on KGs. Recently, CompGCN \cite{DBLP:conf/iclr/VashishthSNT20} extends this idea and leverages a variety of composition operations between KG entities and relations. It shows great effectiveness on KG reasoning tasks.

\paragraph{Temporal knowledge graph embedding models.} 
Temporal knowledge graph embedding models can be categorized into several classes according to their temporal information encoding techniques. A series of models treat every timestamp separately and assign a high-dimensional vector as its embedding \cite{DBLP:conf/esws/TrespMBY17,DBLP:conf/www/LeblayC18,DBLP:conf/iclr/LacroixOU20}. The assigned timestamp embeddings lie in the same space as entity and relation embeddings. Another series of models assume that every entity has a time-aware embedding that evolves over time \cite{DBLP:conf/semweb/XuNAYL20,DBLP:conf/aaai/GoelKBP20}. To achieve time-aware property, an entity together with a timestamp are input into a function (or neural network) to yield a time-aware entity representation at this timestamp. Besides, Garc\'{i}a-Dur\'{a}n et al. jointly encode entity, relation and time information with \textit{Recurrent Neural Network} (RNN) to learn time-aware graph representations \cite{DBLP:conf/emnlp/Garcia-DuranDN18}. Some recent models attempt to model time difference, i.e., time displacement, between the query event and known events. It turns out that time displacement modeling can contribute to superior performance on TKG reasoning tasks, including TKGC \cite{DBLP:conf/emnlp/WuCCH20,DBLP:conf/kdd/JungJK21} and TKG few-shot learning \cite{ding2022few,DBLP:journals/corr/abs-2205-10621}.

\paragraph{Temporal knowledge graph completion.} 
Let $\mathcal{E}$, $\mathcal{R}$ and $\mathcal{T}$ denote a finite set of entities, relations and timestamps, respectively. A temporal knowledge graph $\mathcal{G}$ is a graph which represents the evolution of interactions among entities over time. At any timestamp $t \in \mathcal{T}$, $\mathcal{G}(t)$ is called the TKG snapshot at $t$, and it can be taken as a static KG containing the facts valid at $t$. Any fact, i.e., event, can be described with a quadruple $(s,r,o,t)$, where $s\in \mathcal{E}$ represents the subject, $o \in \mathcal{E}$ represents the object, $r \in \mathcal{R}$ represents the relation between $s$ and $o$, and $t \in \mathcal{T}$ indicates the timestamp when this fact is valid. Therefore, at $t$, the TKG snapshot can be summarized as a finite set of all the valid facts at this timestamp $t$, i.e., $\mathcal{G}(t) = \{(s,r,o,t)|s, o\in \mathcal{E}, r \in \mathcal{R}\}$. We denote a TKG as a sequence of TKG snapshots $\mathcal{G} = \{\mathcal{G}(1), ..., \mathcal{G}(T)\}$, where $T = |\mathcal{T}|$ is the number of timestamps. Similarly, we can also denote a TKG as a finite set of all valid facts which happen at any timestamp $t \in \mathcal{T}$, i.e., $\mathcal{G} = \{(s,r,o,t)|s, o\in \mathcal{E}, r \in \mathcal{R}, t \in \mathcal{T}\}$.
We define the TKGC task as follows. For every snapshot $\mathcal{G}(t)$ in an \textit{observed} TKG $\mathcal{G} = \{\mathcal{G}(1), ..., \mathcal{G}(T)\}$, it contains all the \textit{observed} facts at $t$. Let $\Bar{\mathcal{G}}(t)$ denote the set of all the \textit{true} facts at $t$ such that $\mathcal{G}(t) \in \Bar{\mathcal{G}}(t)$. TKGC aims to predict the ground truth object (or subject) entities of queries $(s, r, ?, t)$ (or $(?, r, o, t)$), where $(s, r, o, t)\in \Bar{\mathcal{G}}(t)$ but $(s, r, o, t)\notin \mathcal{G}(t)$, given any $t\in \mathcal{T}$. TKGC has recently gained increasing interest. Researchers have paid great attention to better modeling the temporal information brought by the nature of TKGs. As fancier techniques and advanced deep learning methods, e.g., attention mechanisms and Transformers \cite{DBLP:conf/nips/VaswaniSPUJGKP17}, being extensively studied, recent TKG reasoning models \cite{DBLP:conf/emnlp/WuCCH20,DBLP:conf/kdd/JungJK21} benefit from them and show great performance on TKGC.

\section{Our method}
\begin{figure}[htbp]
\centering
\includegraphics[width=0.5\columnwidth]{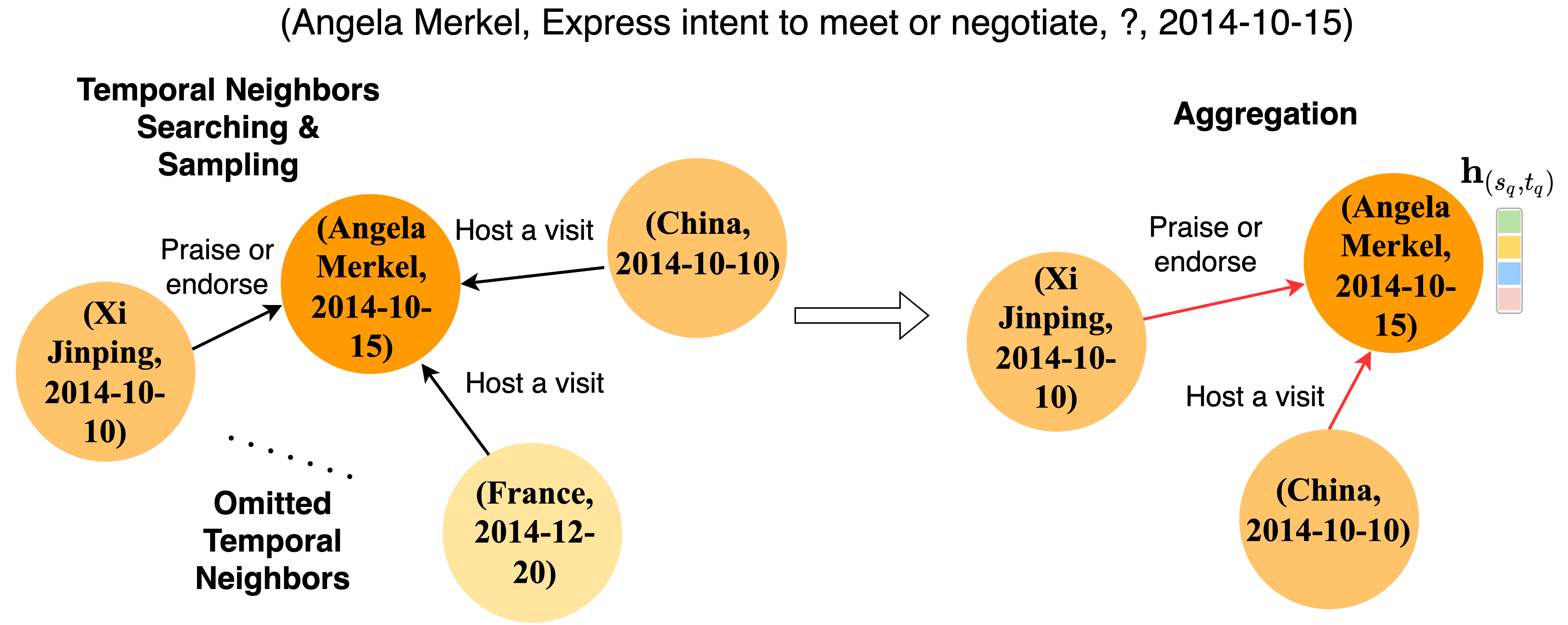}
\caption{\label{fig: overview}The encoding process in TARGCN for the query (\textit{Angela Merkel}, \textit{Express intent to meet or negotiate}, ?, \textit{2014-10-15}). The color darkness on each node implies its probability of being sampled as an input at the aggregation step (the darker the higher).}
\end{figure}

To solve the TKGC task, our relational graph encoder TARGCN extensively collects information from the whole temporal context and updates the time-aware representations of entities. For every link prediction query $(s_q,r_q,?,t_q)$, TARGCN first creates a subgraph for the subject $s_q$, according to its temporal associated neighbors. Then it derives time-aware representations for the neighbors from the temporal neighborhood, and performs aggregation. After $s_q$'s time-aware representation is updated, a knowledge graph decoder (score function) is utilized to compute scores for every candidate object, which yields the plausibility of every candidate object being the ground truth object in the link prediction query $(s_q,r_q,?,t_q)$. Note that we only consider object prediction queries $(s_q,r_q,?,t_q)$ in our work since we add reciprocal relations
for every quadruple , i.e., adding $(o,r^{-1},s,t)$ for
every $(s,r,o,t)$. The restriction to only predict
object entities does not lead to a loss of generality. An example is presented in Figure \ref{fig: overview} which shows the encoding process of our model. For the query subject \textit{Angela Merkel} appearing at \textit{2014-10-15}, TARGCN selects its temporal neighbors with a time difference-dependent probability. Node aggregation is then performed to learn a contextualized representation $\mathbf{h}_{(s_q,t_q)}$, where $s_q$, $t_q$ correspond to \textit{Angela Merkel} and \textit{2014-10-15}, respectively.


\subsection{Subgraph sampling in temporal neighborhood}
Given a TKGC query $(s_q,r_q,?,t_q)$, TARGCN aims to learn a contextualized representation for the subject entity $s_q$. Inspired by the inference graph proposed in \cite{DBLP:conf/iclr/HanCMT21}, we sample a \textit{Temporal Neighboring Graph} (TNG) for $(s_q, t_q)$ in TKGC context, where $(s_q, t_q)$ is the node representing $s_q$ at $t_q$. We first find out all the temporal neighbors of $(s_q, t_q)$, which can be described as a set  
$\mathcal{N}_{(s_q,t_q)} = \{(e,t)|(e, r, s_q, t) \in \mathcal{G}; e\in \mathcal{E}, t\in \mathcal{T},r \in \mathcal{R}\}$. The entity $e$ of a temporal neighbor $(e,t)$ forms a link with $s_q$ at timestamp $t$ and $s_q$ bears an incoming edge derived from the temporal associated quadruple $(e, r, s_q, t)$. Note that in TKGC, though we cannot observe all the true quadruples, we still can observe part of true quadruples at \textit{every} timestamp. This enables TARGCN to search for the temporal neighbors of $(s_q, t_q)$ along the whole time axis. Then we employ weighted sampling strategy according to the absolute time difference $|t_q-t|$ between $(s_q, t_q)$ and the corresponding temporal neighbor $(e,t)$. For every temporal neighbor $(e,t)$, the probability of it being sampled into $(s_q, t_q)$'s TNG is computed by: $exp(-|t_q-t|)/\Sigma_{(e, t')\in \mathcal{N}_{(s_q,t_q)}}exp(-|t_q-t'|)$. In this way, higher probabilities are assigned to the temporal neighbors who are closer to $(s_q, t_q)$ along the time axis. We adopt this sampling strategy since we assume that for the inference of a fact at $t_q$, it is more likely to find clues from the factual information at nearer timestamps. Besides, we use a hyperparameter to limit the maximum number of the temporal neighbors included in $(s_q, t_q)$'s TNG to prevent over sampling \textit{less-concerned} temporal neighbors. An example illustrating $(s_q,t_q)$'s TNG is shown in Figure \ref{fig: TGN}. In the process of TNG sampling, TARGCN does not include any parameter. For $(s_q, t_q)$, TARGCN selects the contributive temporal neighbors and generates a comprehensive temporal context of it with a parameter-free TNG sampler, rather than using a trainable component, e.g., a self-attention-based module employed in recent neural-based TKGC methods \cite{DBLP:conf/emnlp/WuCCH20,DBLP:conf/kdd/JungJK21}.


\subsection{Time-aware relational aggregation}
After sampling TNG for the subject entity $s_q$, we then attempt to learn its contextualized representation through neighborhood aggregation. Since we have access to temporal neighbors from the whole timeline, we implicitly incorporate temporal information. Inspired by \cite{DBLP:conf/iclr/XuRKKA20}, we employ a functional time encoder for reasoning TKGs, and learn a time-aware entity representation for every temporal neighbor. In this way, we are able to distinguish the temporal neighbors, $(e,t)$ and $(e,t')$, who root from the same entity $e$ but emerge at different timestamps $t$ and $t'$. The time-aware entity representation is computed as:
\begin{equation}
    \mathbf{h}_{(e,t)} = f(\mathbf{h}_e \| \boldsymbol{\Phi}(t, t_q)),
\end{equation}
where $\mathbf{h}_e \in \mathbb{R}^{d_e}$ denotes the time-invariant entity-specific representation of the entity $e$. $\boldsymbol{\Phi}(t, t_q) = \sqrt{\frac{1}{d_t}}[cos(\omega_1(t-t_q)+\phi_1), ..., cos(\omega_{d_t}(t-t_q)+ \phi_{d_t}))]$ is a time difference encoder mapping $t-t_q$ to a finite dimensional functional space $\mathbb{R}^{d_t}$, where $\omega_1$ to $\omega_{d_t}$ are trainable frequency components, $\phi_1$ to $\phi_{d_t}$ are trainable phase components. We concatenate the time-invariant entity representation with its corresponding time difference representation, and learn a combined representation of them with a layer of feed-forward neural network $f$. Note that the sign of $t-t_q$ will affect the output of the time difference encoding module. We aggregate the information from $(s_q, t_q)$'s temporal neighbors with a relational graph aggregator:
\begin{equation}
    \mathbf{h}_{(s_q,t_q)} = \frac{1}{|\Bar{\mathcal{N}}_{(s_q,t_q)}|} \sum_{(e,t) \in \Bar{\mathcal{N}}_{(s_q,t_q)}}\mathbf{W}(\mathbf{h}_{(e,t)}\|\mathbf{h}_r).
\end{equation}
$\Bar{\mathcal{N}}_{(s_q,t_q)}$ denotes a finite set of temporal neighbors sampled from $
(s_q, t_q)$'s temporal neighborhood, i.e., all the neighbors in $(s_q, t_q)$'s TNG. $r$ is the relation appearing in the temporal associated quadruple $(e, r, s_q, t)$ where temporal neighbor $(e, t)$ is sampled. We assume that relation representations are time-invariant and we incorporate relational information into the graph encoder by concatenating time-aware node representations with them. Our graph encoder outputs the time-aware representation of $s_q$ at query time $t_q$, by combining not only the raw entity representation $\mathbf{h}_e$ but also the implicit time difference information from its temporal neighbors.
\begin{figure}[htbp]
\centering
\includegraphics[width=0.6\columnwidth]{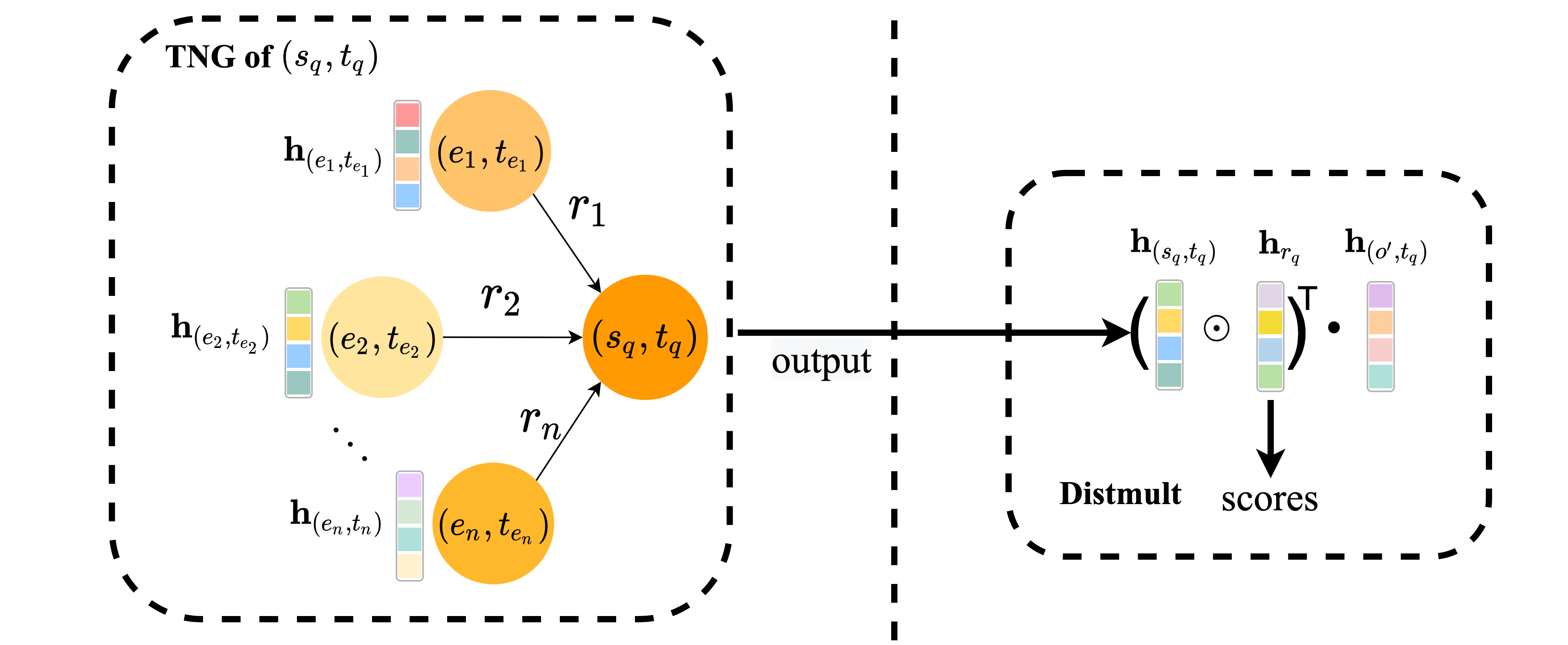}
\caption{\label{fig: Inference} Inference process of TARGCN + Distmult. $\mathbf{h}_{(o',t_q})$ is the time-aware representation of a candidate $o'$ at $t_q$. For a TKGC query $(s_q,r_q,?, t_q)$, we first sample a TNG rooting from $(s_q, t_q)$. Then we employ TARGCN encoder to compute the representation $\mathbf{h}_{(s_q,t_q)}$ for $(s_q,t_q)$. We provide Distmult with time-aware representations of all candidates for score computation. The candidate producing the highest score is selected as the predicted answer.}
\end{figure}

\subsection{Learning and inference}
Figure \ref{fig: Inference} illustrates how TARGCN, together with a KG score function, i.e., Distmult \cite{DBLP:journals/corr/YangYHGD14a}, predicts the ground truth missing object for the TKGC query $(s_q,r_q,?, t_q)$. Given $s_q$, we use the sampling strategy and our time-aware relational graph encoder to compute a time dependent node representation for $(s_q,t_q)$. Then we use a KG score function to compute the plausibility of every candidate entity. TARGCN can be coupled with any KG score function. We choose TARGCN + Distmult as the final model structure because it achieves a high parameter efficiency compared with TARGCN coupled with other two KG score functions, i.e., ComplEx \cite{DBLP:conf/icml/TrouillonWRGB16} and BiQUE \cite{guo-kok-2021-bique}, on the benchmark datasets (discussed in Appendix \ref{sec: appendix kg score}), which encounters our flavor of building a parameter-eficient TKGC model. Note that for the candidate entities, we do not sample TNG for them to avoid huge time consumption during inference. Instead, for every candidate entity $o'$, we simply derive its time-aware representation by computing $\mathbf{h}_{(o',t_q)} = f(\mathbf{h}_{o'} \| \boldsymbol{\Phi}(t_q, t_q))$. The temporal encoder  $\boldsymbol{\Phi}(\cdot, \cdot)$ will also return a unique representation when time difference equals zero. We employ cross-entropy loss for parameter learning:
\begin{equation}
\resizebox{.5\textwidth}{!}{$
    \mathcal{L} =  \sum \limits_{(s, r, o, t)\in \mathcal{G}} 
    -\textrm{log}
    \left(\frac{score(\mathbf{h}_{(s,t)},\mathbf{h}_r,\mathbf{h}_{(o,t)})}{\Sigma_{o' \in \mathcal{E}}score(\mathbf{h}_{(s,t)},\mathbf{h}_r,\mathbf{h}_{(o',t)})}\right)$}
    ,
\end{equation}
where $o'$ denotes all candidate entities and we sum over all \textit{observed} quadruples in $\mathcal{G}$. Note that our TARGCN encoder can be equipped with any KG score functions since our encoder returns time-aware representations for entities. In our work, $score(\mathbf{h}_{(s,t)},\mathbf{h}_r,\mathbf{h}_{(o',t)}) = (\mathbf{h}_{(s,t)} \odot \mathbf{h}_r)^\top\mathbf{h}_{(o',t)}$, where $\odot$ denotes the Hadamard product.

\section{Experiments}
\label{sec: experiment}
We compare our model with several existing TKGC methods on three TKGC benchmark datasets. We prove the robustness of TARGCN and present ablation studies. To show the parameter efficiency of our model, we further do an analysis of parameter usage on TARGCN.

\subsection{Experimental setup}
\paragraph{Datasets.} 
We perform evaluation on three TKGC benchmark datasets: (1) ICEWS14 \cite{DBLP:conf/emnlp/Garcia-DuranDN18} (2) ICEWS05-15 \cite{DBLP:conf/emnlp/Garcia-DuranDN18} (3) GDELT \cite{leetaru2013gdelt}. ICEWS14 and ICEWS05-15 are two subsets of \textit{Integrated Crisis Early Warning System} (ICEWS) database. ICEWS14 contains timestamped political facts happening in 2014, while the timestamps of factual events in ICEWS05-15 span from 2005 to 2015. We follow \cite{DBLP:conf/emnlp/WuCCH20} and use the GDELT subset proposed by \cite{DBLP:conf/icml/TrivediDWS17}. It contains global social facts from April 1, 2015 to March 31, 2016. The detailed dataset statistics are presented in Table \ref{tab: data} in Appendix \ref{sec:appendix b}.

\paragraph{Evaluation metrics.} 
We employ two evaluation metrics for all experiments, i.e., Hits@1/3/10 and \textit{Mean Reciprocal Rank} (MRR). For every test fact $(s_q, r_q, o_q, t_q) \in \Bar{\mathcal{G}}$ ($(s_q, r_q, o_q, t_q) \notin \mathcal{G}$), we derive an associated TKGC query $q = (s_q, r_q,?,t_q)$. We let models compute the rank $\psi$ of the ground truth entity $o_q$ among all the candidates. Hits@1/3/10 are the proportions of the test facts where ground truth entities are ranked as top 1, top 3, top 10, respectively. MRR is defined as $\frac{1}{|\mathcal{Q}|}\sum_{q \in \mathcal{Q}} \frac{1}{\psi}$, where $\mathcal{Q}$ denotes the set of all queries. It computes the mean of the reciprocal ranks of ground truth entities. We follow the \textit{filtered} setting proposed by \cite{DBLP:conf/nips/BordesUGWY13} to achieve fairer evaluation.

\paragraph{Baseline methods.} 
We take ten methods as baseline models. The first four baselines are static KG reasoning methods, i.e., 
ComplEx \cite{DBLP:conf/icml/TrouillonWRGB16} and SimplE \cite{DBLP:conf/nips/Kazemi018}. The other methods are developed to solve TKGC, including 
DE-SimplE \cite{DBLP:conf/aaai/GoelKBP20}, ATiSE \cite{DBLP:conf/semweb/XuNAYL20}, TNTComplEx \cite{DBLP:conf/iclr/LacroixOU20}, ChronoR \cite{DBLP:conf/aaai/SadeghianACW21}, TeLM \cite{DBLP:conf/naacl/XuCNL21}, BoxTE \cite{DBLP:journals/corr/abs-2109-08970}, TeMP \cite{DBLP:conf/emnlp/WuCCH20} and T-GAP \cite{DBLP:conf/kdd/JungJK21}. Among all baselines, only TeMP and T-GAP are neural-based methods that employ GNNs as graph encoders, similar to our TARGCN setting. Therefore, we further compare the parameter efficiency among them.
\subsection{Experimental results}
\label{sec: exp results}
\begin{table}[t]
\caption{Temporal knowledge graph completion results on three benchmark datasets. Evaluation metrics are
    filtered MRR and Hits@1/3/10.
    The best results are marked in bold. Results marked with $[\blacktriangledown]$, $[\heartsuit]$, $[\bigstar]$ are taken from \protect\cite{DBLP:conf/emnlp/WuCCH20}, \protect\cite{DBLP:conf/kdd/JungJK21}, \protect\cite{DBLP:journals/corr/abs-2109-08970}, respectively.}
\label{tab: tid link prediction results}

    \centering
    \resizebox{0.77\textwidth}{!}{
    \large\begin{tabular}{@{}lcccccccccccc@{}}
\toprule
        \textbf{Datasets} & \multicolumn{4}{c}{\textbf{ICEWS14}} &  \multicolumn{4}{c}{\textbf{ICEWS05-15}} & \multicolumn{4}{c}{\textbf{GDELT}}\\
\cmidrule(lr){2-5} \cmidrule(lr){6-9} \cmidrule(lr){10-13}
        \textbf{Model} & MRR & Hits@1 & Hits@3 & Hits@10 & MRR & Hits@1 & Hits@3  & Hits@10 & MRR & Hits@1 & Hits@3  & Hits@10 \\
\midrule 
        ComplEx $[\blacktriangledown]$ & 0.442 & 0.400 & 0.430 & 0.664
        & 0.464 & 0.347 & 0.524 & 0.696 
        & 0.213 & 0.133 & 0.225 & 0.366
         \\
        SimplE $[\blacktriangledown]$ & 0.458 & 0.341 & 0.516 & 0.687
        & 0.478 & 0.359 & 0.539 & 0.708 
        & 0.206 & 0.124 & 0.220 & 0.366
         \\
\midrule
       
       
       DE-SimplE $[\blacktriangledown]$ & 0.526 & 0.418 & 0.592 & 0.725
       & 0.513 & 0.392 & 0.578 & 0.748
       & 0.230 & 0.141 & 0.248 & 0.403 
        \\
       
       ATiSE $[\blacktriangledown]$ & 0.571 & 0.465 & 0.643 & 0.755
       & 0.484 & 0.350 & 0.558 & 0.749
       & - & - & - & -
        \\
       
       TNTComplEx $[\blacktriangledown]$ & 0.620 & 0.520 & 0.660 & 0.760 
       & 0.670 & 0.590 & 0.710 & 0.810 
       & - & - & - & - 
       \\
       ChronoR $[\bigstar]$ & 0.625 & 0.547 & 0.669 & 0.773 
       & 0.675 & 0.596 & 0.723 & 0.820 
       & - & - & - & - 
       \\
       TeLM $[\bigstar]$ & 0.625 & 0.545 & 0.673 & 0.774 
       & 0.678 & 0.599 & 0.728 & 0.823 
       & - & - & - & - 
       \\
       BoxTE $[\bigstar]$ & 0.613 & 0.528 & 0.664 & 0.763 
       & 0.667 & 0.582 & 0.719 & 0.820 
       & 0.352 & 0.269 & 0.377 & 0.511 
       \\
\midrule
        TeMP-GRU $[\blacktriangledown]$
        & 0.601 & 0.478 & 0.681 & 0.828
        & 0.691  & 0.566 & \textbf{0.782} & \textbf{0.917} 
        & 0.275 & 0.191 & 0.297 & 0.437  
        \\
        TeMP-SA $[\blacktriangledown]$
        & 0.607 & 0.484 & \textbf{0.684} & \textbf{0.840}
        & 0.680  & 0.553 & 0.769 & 0.913
        & 0.232 & 0.152 & 0.245 & 0.377  
        \\
        T-GAP $[\heartsuit]$
        & 0.610 & 0.509 & 0.677 & 0.790
        & 0.670  & 0.568 & 0.743 & 0.845
        & - & - & - & -  
        \\
\midrule
        TARGCN 
        & \textbf{0.636} & \textbf{0.576} & 0.672 & 0.746
        & \textbf{0.702}  & \textbf{0.635} & 0.743 & 0.823
        & \textbf{0.515} & \textbf{0.423} & \textbf{0.557} & \textbf{0.689}  
        \\
        & $\pm$ 0.001 & $\pm$ 0.003 & $\pm$ 0.001 & $\pm$ 0.003
        & $\pm$ 0.001 & $\pm$ 0.003 & $\pm$ 0.002 & $\pm$ 0.002
        & $\pm$ 0.002 & $\pm$ 0.002 & $\pm$ 0.001 & $\pm$ 0.003
        \\
\bottomrule
    \end{tabular}
    }
\end{table}
Table \ref{tab: tid link prediction results} reports the experimental results of all methods on three benchmark datasets. We can observe that TARGCN outperforms all baselines on all datasets. The margin is particularly huge on the GDELT dataset. TARGCN achieves an over 46\% relative improvement on MRR compared with the strongest baseline BoxTE. TARGCN also leads in Hits metrics greatly. It improves Hits@1/3/10 by 57.25\%, 47.75\%, and 34.83\%, respectively. On ICEWS datasets, TARGCN still achieves the best results on MRR and Hits@1. We argue that the performance gap varies because of the characteristics of different datasets. While ICEWS datasets are sparse, GDELT is much denser. As discussed in \cite{DBLP:conf/emnlp/WuCCH20,DBLP:journals/corr/abs-2109-08970}, the temporal sparsity issue on ICEWS is much more severe than it on GDELT. This implies that GDELT contains substantially more temporal patterns, while ICEWS datasets are more prone to be biased by a large number of isolated events which are mainly dominated by sparse entities and relations. Hence, we argue that reasoning on GDELT requires much stronger techniques. 
For prior methods, though several TKGC methods outperform static methods on GDELT, the improvements are not substantial. However, TARGCN achieves a more than 141\% relative improvement on MRR, compared with the strongest static KG baseline ComplEx. This shows the superior effectiveness of our graph encoder in capturing various temporal patterns. For ICEWS datasets, our model can also achieve state-of-the-art performance. This demonstrates its strong ability in capturing the temporal KG information brought by sparse entities and relations.

\subsection{Model analysis}
\label{sec: model analysis}
\begin{table}[htbp]
\caption{Performance of generalization to unseen timestamps and irregular timestamped data.
    }\label{tab: unseen}
    \centering
    \resizebox{0.5\columnwidth}{!}{
    \large\begin{tabular}{@{}lcccccccc@{}}
\toprule
        \textbf{Datasets} & \multicolumn{4}{c}{\textbf{ICEWS14-unseen}} &  \multicolumn{4}{c}{\textbf{ICEWS14-irregular}}\\
\cmidrule(lr){2-5} \cmidrule(lr){6-9}
        \textbf{Model} & MRR & Hits@1 & Hits@3 & Hits@10 & MRR & Hits@1 & Hits@3 & Hits@10  \\
\midrule 
        TComplEx  & 0.461 & 0.365 & 0.513 & 0.644 & 0.509 & 0.421 & 0.558 & 0.678
         \\
        TNTComplEx  & 0.474 & 0.373 & 0.524 & 0.665 & 0.512 & 0.429 & 0.558 & 0.665
         \\
        TeMP-SA & - & - & - & - & 0.521 & 0.408 & 0.583 & \textbf{0.741}
         \\
        T-GAP  & 0.474 & 0.362 & 0.532 & 0.689 & 0.526 & 0.428 & \textbf{0.588} & 0.719 
         \\
        TARGCN & \textbf{0.578} & \textbf{0.518} & \textbf{0.607} & \textbf{0.692} & \textbf{0.552} & \textbf{0.496} & 0.583 & 0.667
        \\

\bottomrule
    \end{tabular}
    }
\end{table}
\paragraph{Generalization to unseen timestamps and irregular timestamped data.} 
To prove the robustness of our model, we follow \cite{DBLP:conf/aaai/GoelKBP20} to test its ability to predict the links at unseen timestamps. We exclude every quadruple appearing on the 5th, 15th, and 25th day of each month in ICEWS14 to construct a new training set. We randomly split the excluded quadruples into validation and test sets. We compare TARGCN with several recently proposed baselines on this new dataset ICEWS14-unseen, and the results (Table \ref{tab: unseen}) indicate the strong robustness of our model on timestamp generalization. TARGCN greatly outperforms all baseline methods, especially in Hits@1. 
To infer links at a timestamp, TeMP requires at least one fact seen at this timestamp during training, thus making it unable to generalize to unseen timestamps. T-GAP employs discretized time displacement embeddings. It trains different embeddings for different time differences. If some time differences appear much fewer times in the training set, the corresponding time displacement embeddings will not be well trained. Compared with these two advanced neural-based TKGC methods, TARGCN not only has the ability to perform link prediction at unseen timestamps, but also shows superior generalization power. TARGCN computes time-aware representations with a functional time encoder which is jointly trained on any time difference seen in the training set, which helps it outperform T-GAP greatly.

Besides, we propose another new dataset ICEWS14-irregular to validate whether TKGC models can generalize well to the TKG data collected at irregular-spaced timestamps. We randomly sample the snapshots in ICEWS14 and keep the time interval between every two of the sampled neighboring snapshots not greater than 4. We perform TKGC on ICEWS14-irregular and experimental results in Table \ref{tab: unseen} show that TARGCN is superior in handling data with irregular timestamps. Compared with TARGCN who takes advantage of the graph information from the whole timeline, TeMP utilizes a fixed short time span of graph information to learn embeddings, which fails to capture a large amount of graph information outside this span. T-GAP uses time displacement embeddings to encode different time differences. However, experimental results show that TARGCN distinguishes irregular time intervals better than T-GAP with the help of its functional time encoder that computes the representation of any time difference with shared parameters. Appendix \ref{sec: appendix e} provides more details.


\begin{figure}[htbp]
\centering
\begin{tikzpicture}[scale=0.4, transform shape]
  \pgfplotsset{width=14cm, height=7.5cm, compat=1.16, legend style={at={(0.63,0.18)},anchor=west}
  }
  \definecolor{clr1}{HTML}{12AAB5}
  \definecolor{clr2}{HTML}{E85642}
  \definecolor{clr3}{HTML}{9933FF}
  \definecolor{clr4}{HTML}{000000}
\begin{axis}[%
  xmode=log,
  xlabel=Number of Parameters (log scale),
  ylabel=Filtered MRR,
  ymajorgrids=true,
  grid style=dashed,
  ymin=0.56, ymax=0.65,
  xmin=50e4,xmax=40e5, 
  scale = 1.03
]
  \addplot+[mark=square*, mark options={fill=clr2}, clr2] coordinates {(799400, 0.605) (1229100,0.627) (1678800,0.629) (2638200, 0.636)};
  \addlegendentry{TARGCN}
  \addplot+[mark=square*, mark options={solid, fill=clr1}, clr1] coordinates {(928675, 0.582)(1912350,0.61)(2951025, 0.596)};
  \addlegendentry{T-GAP}
  \addplot+[mark=square*, mark options={solid, fill=clr4}, clr4] coordinates {(611840,0.595) (1264640,0.607) (2928640, 0.618)};
  \addlegendentry{TeMP-SA}
\end{axis}
\end{tikzpicture}
\caption{Filtered MRR on ICEWS14 achieved by TARGCN, T-GAP and TeMP-SA, with varied number of parameters. More details in Appendix \ref{sec: appendix param eff}.
}
\label{fig: efficiency 14}
\end{figure}
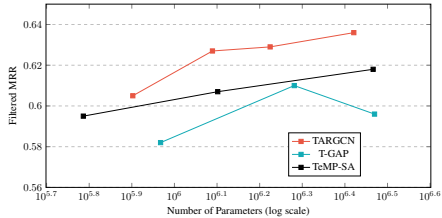

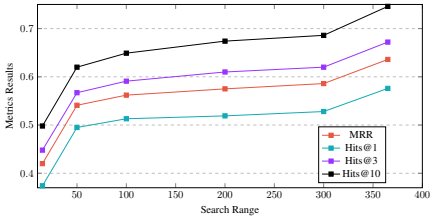
\begin{figure}[htbp]
\centering
\begin{tikzpicture}[scale=0.4, transform shape]
  \pgfplotsset{width=14cm, height=7.5cm, compat=1.16, legend style={at={(0.73,0.18)},anchor=west}}
  \definecolor{clr1}{HTML}{12AAB5}
  \definecolor{clr2}{HTML}{E85642}
  \definecolor{clr3}{HTML}{9933FF}
  \definecolor{clr4}{HTML}{000000}
\begin{axis}[%
  xlabel=Search Range,
  ylabel=Metrics Results,
  ymajorgrids=true,
  grid style=dashed,
  ymin=0.37, ymax=0.75,
  xmin=10,xmax=400, 
  scale = 1.03
]
  \addplot+[mark=square*, mark options={fill=clr2}, clr2] coordinates {(15, 0.420) (50,0.541) (100,0.562) (200, 0.575)(300, 0.586) (365, 0.636)};
  \addlegendentry{MRR}
  \addplot+[mark=square*, mark options={solid, fill=clr1}, clr1] coordinates {(15,0.374)(50,0.495)(100, 0.513)(200,0.519)(300,0.528)(365, 0.576)};
  \addlegendentry{Hits@1}
  \addplot+[mark=square*, mark options={solid, fill=clr3}, clr3] coordinates {(15,0.448)(50,0.567)(100, 0.591)(200,0.610)(300,0.620)(365,0.672)};
  \addlegendentry{Hits@3}
  \addplot+[mark=square*, mark options={solid, fill=clr4}, clr4] coordinates {(15,0.498)(50,0.620)(100, 0.649)(200,0.674)(300,0.686)(365,0.746)};
  \addlegendentry{Hits@10}
\end{axis}
\end{tikzpicture}
\caption{Temporal neighborhood exploration analysis on ICEWS14.}
\label{fig: temporal neigh exp}
\end{figure}

\begin{table}[htbp]
\caption{Parameter efficiency comparison on ICEWS05-15 and GDELT. Due to extremely high memory consumption, we cannot train T-GAP on GDELT even when batch size equals 1.}
\label{tab: efficiency GDELT,ICEWS05-15}
    \centering
    \resizebox{0.58\columnwidth}{!}{
    \small\begin{tabular}{@{}lcccccccc@{}}
\toprule
        \textbf{Datasets} & \multicolumn{4}{c}{\textbf{ICEWS05-15}} & \multicolumn{4}{c}{\textbf{GDELT}}\\
\cmidrule(lr){2-5} \cmidrule(lr){6-9}
        \textbf{Model} &  Parameters & MRR & Parameter $\uparrow$ & MRR $\downarrow$ &  Parameters & MRR & Parameter $\uparrow$ & MRR $\downarrow$ \\
\midrule 
        TARGCN  & 2359200 & 0.702 & - & - & 269200 & 0.515 & - & -
         \\
        T-GAP  & 3088000 & 0.670 & 30.89\% & 4.56\% & - & - & - & - 
         \\
        TeMP-SA & 2645760 & 0.680 & 12.15\% & 3.13\% & 255232 & 0.232 & -5.19\% & 54.95\%
         \\
        TeMP-GRU  & 2794528 & 0.691 &18.45\% & 1.57\% &404000 & 0.275 & 50.07 \% & 46.60\%
         \\

\bottomrule
    \end{tabular}
    }
    
\end{table}
\paragraph{Parameter efficiency analysis.} 
While TARGCN serves as a strong TKGC model, it also keeps a low parameter cost. We compare the parameter efficiency among TARGCN and two recently proposed neural-based TKGC models, i.e., TeMP and T-GAP. 
On ICEWS14, for all three models, we adjust the embedding size of both entities and relations to adjust the number of parameters. We do not change model structures and other hyperparameter settings. In Figure \ref{fig: efficiency 14}, we show that TARGCN performs better as we increase model parameters. More importantly, even with much fewer parameters, TARGCN still outperforms TeMP and T-GAP.
For ICEWS05-15 and GDELT, we summarize the number of parameters as well as performance difference in Table \ref{tab: efficiency GDELT,ICEWS05-15}. We compare across the models with parameter settings that lead to the experimental results shown in Table \ref{tab: tid link prediction results}. We show that TARGCN enjoys superior parameter efficiency, especially on GDELT.
We attribute such high parameter efficiency to our simple but powerful graph encoder. Note that in the TNG sampling process, we force our model to choose the temporal neighbors who are nearer to the source node $(s_q, t_q)$ on the time axis, by assigning higher sampling probabilities to them.
Models like TeMP and T-GAP employ self-attention modules to let models choose their attention themselves through parameter learning. We argue that even if such modules are powerful, they can be simplified in the context of TKGC. In our model, we force our TNG sampler to focus on the facts happening at the timestamps that are closer to the query timestamp. Our TNG sampling process does not include any parameter, while self-attention modules increase parameters, cause higher memory consumption, and bring heavier burdens for parameter optimization. Apart from that, compared with TeMP who encodes temporal information only from a fixed short time span of $2\tau$, our TNG sampling range spans across the whole timeline. This means that even if a temporal neighbor is derived from a sparse entity and it appears only at faraway timestamps from the query timestamp, our sampler still has the ability to include it into the TNG and enable information aggregation. Similar to TARGCN, T-GAP, with the help of its \textit{Preliminary GNN} (PGNN), is able to find any temporal associated quadruples related to any entity appearing at any time. However, in its PGNN, it employs three weight matrices together with discretized time displacement embeddings $\mathbf{h}_{|\Delta t|}$ to fully express the supporting information coming from the past, the present and the future. We find it redundant to model time difference in this way. 
In TARGCN, we do not use separate weight matrices during aggregation since our functional time encoder distinguishes the sign of time difference itself.
Besides, instead of learning different discretized embeddings to represent different $|\Delta t|$, our model computes the representation of any time difference with shared parameters, thus cutting parameter consumption.
\paragraph{Ablation study.}
To validate the effectiveness of different model components, we conduct several ablation studies on ICEWS14 and GDELT. We first change the time difference encoding module into an absolute time encoder, e.g., for a $(s_q, t_q)$ and a temporal neighbor $(e, t)$, we learn a representation for $t$ instead of $t-t_q$. From Table \ref{tab: ablation}, we observe performance drops on both datasets. This proves the effectiveness of time difference modeling. Next, we adopt random sample in TNG sampling process.
The performance drops on both datasets, indicating that by sampling more neighbors nearer in the temporal context, our model benefits more in learning better representations. Additionally, we conduct another experiment by including all temporal neighbors during aggregation. We observe huge performance drops on both datasets, which proves that our sampling strategy helps to exclude noisy information from less-concerned neighbors.
\begin{table}[htbp]
\caption{Ablation studies of TARGCN variants on ICEWS14 and GDELT.
    }
    \label{tab: ablation}
    \centering
    \resizebox{0.55\columnwidth}{!}{
    \large\begin{tabular}{@{}lcccccccc@{}}
\toprule
        \textbf{Datasets} & \multicolumn{4}{c}{\textbf{ICEWS14}} &  \multicolumn{4}{c}{\textbf{GDELT}}\\
\cmidrule(lr){2-5} \cmidrule(lr){6-9}
        \textbf{Model} & MRR & Hits@1 & Hits@3 & Hits@10 & MRR & Hits@1 & Hits@3 & Hits@10  \\
\midrule 
        Absolute Time  & 0.622 & 0.556 & 0.660 & 0.739 & 0.502 & 0.408 & 0.545 & 0.678
         \\
        Random Sample  & 0.618 & 0.551 & 0.656 & 0.735 & 0.433 & 0.312 & 0.502 & 0.640
         \\
        Whole Neighborhood & 0.481 & 0.433 & 0.501 & 0.568 & 0.431 & 0.312 & 0.497 & 0.633
         \\
        TARGCN &  \textbf{0.636} & \textbf{0.576} & \textbf{0.672} & \textbf{0.746}
        & \textbf{0.515} & \textbf{0.423} & \textbf{0.557} & \textbf{0.689}
        \\
\bottomrule
    \end{tabular}
    }
    
\end{table}
\paragraph{Temporal neighborhood exploration.}
We further conduct an experiment to study how TARGCN performs while the search range varies. We report in Figure \ref{fig: temporal neigh exp} our model's performance on ICEWS14 with different search range, namely, 15, 50, 100, 200, 300, and 365 (whole timeline). For all the metrics, TARGCN's performance improves greatly and constantly as the search range increases. This proves that the effectiveness of TARGCN mainly comes from its superiority in exploring the temporal context. The amount of available temporal information is decisive for our simple-structured model. Compared with the models that only make use of graph snapshots near to the query timestamp $t_q$, e.g., TeMP, we simplify the model structure but take advantage of as much temporal information as we can.
\section{Conclusion}
We propose a simple but powerful graph encoder TARGCN for \textit{Temporal Knowledge Graph Completion} (TKGC). TARGCN employs a \textit{Temporal Neighboring Graph} (TNG) sampling strategy, which enables it to extensively utilize the information from the whole temporal context. Experimental results show that TARGCN achieves state-of-the-art performance on three benchmark TKGC datasets. 
Besides, TARGCN enjoys a high parameter efficiency. It beats two recently proposed neural-based TKGC methods, i.e., TeMP and T-GAP, with much fewer parameters. Thanks to its time difference learning module and temporal neighbor sampler, TARGCN also shows strong robustness to inferring links on irregular timestamped data or at unseen timestamps. We find that it is not always necessary to incorporate complex modules, e.g., Transformers, into TKG reasoning models. Instead, developing methods to better and more efficiently capture temporal information is more beneficial to TKGC.


\bibliography{sample}
\bibliographystyle{plainnat}

\appendix
\section*{Appendix}

\appendix

\section{Example of temporal neighborhood}
\label{sec:appendix a}
\begin{figure*}[htbp]
\centering
\includegraphics[scale=0.4]{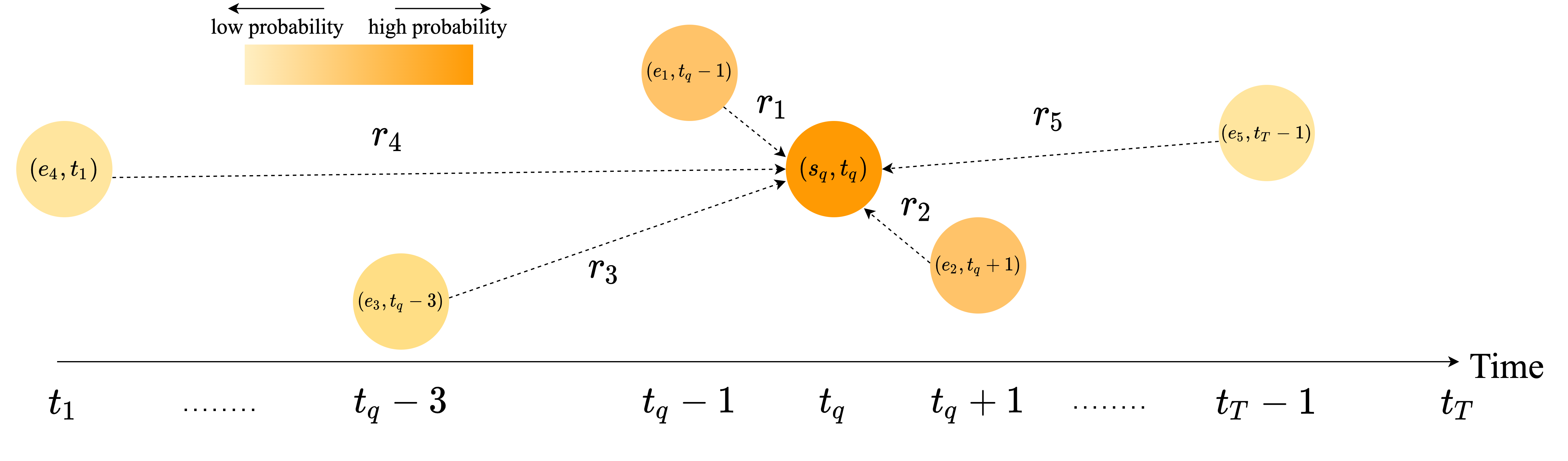}
\caption{\label{fig: TGN} Temporal neighborhood of $(s_q, t_q)$ derived from an object prediction query $(s_q,r_q,?,t_q)$. We use a dashed line (labeled with relation type) to denote a temporal associated link connecting $s_q$ with its temporal neighbor, e.g., the dashed line labeled with $r_4$ corresponds to the temporal associated quadruple $(e_4, r_4, s_q, t_1)$. A temporal neighbor with darker color is assigned a higher probability to be sampled into $(s_q, t_q)$'s TNG. Since $(e_1, t_q-1)$ and $(e_2, t_q+1)$ has the same temporal distance from $t_q$, they are assigned with the same sampling probability (denoted with the same color darkness).}
\end{figure*}

Figure \ref{fig: TGN} shows an example of the temporal neighborhood of $(s_q, t_q)$, generated from a TKGC query $(s_q, r_q, ?, t_q)$. We can represent it as $\mathcal{N}_{(s_q,t_q)} = \{(e_1, t_q-1),(e_2, t_q+1),(e_3, t_q-3), (e_4,t_1),(e_5,t_T-1)\}$. The probability of each temporal neighbor being sampled into $(s_q, t_q)$'s TNG is determined according to the time difference between $t_q$ and the timestamp of this temporal neighbor (the darker the temporal neighbor shows, the higher the probability).

\section{Dataset statistics}
\label{sec:appendix b}

Table \ref{tab: data} contains the dataset statistics of all three benchmark datasets and two newly created datasets, i.e., ICEWS14-unseen and ICEWS14-irregular. The data creation process of ICEWS14-unseen and ICEWS14-irregular is discussed in Appendix \ref{sec: appendix e}.

\begin{table}[htbp]
\caption{Dataset statistics. $N_\textrm{train}$, $N_\textrm{valid}$, $N_\textrm{test}$ represent the number of quadruples in the training set, validation set, and test set, respectively. $|\mathcal{T}|$ denotes the number of timestamps, where we take a snapshot of a TKG at each timestamp. All facts in all datasets are denoted in English.}
\label{tab: data}
    \centering
    \resizebox{0.8\columnwidth}{!}{
\begin{tabular}{c c c c c c c c} \hline
Dataset&$N_\textrm{train}$&$N_\textrm{valid}$&$N_\textrm{test}$&$|\mathcal{E}|$&$|\mathcal R|$&$|\mathcal{T}|$\\ \hline
ICEWS14  & $72,826$ & $8,941$ & $8,963$ & $7,128$ & $230$ & $365$\\ 
ICEWS05-15  & $386,962$ & $46,275$ & $46,092$ & $10,488$ & $251$ & $4,017$\\ 
GDELT  & $2,735,685$ & $341,961$ & $341,961$ & $500$ & $20$ & $366$\\ 
ICEWS14-unseen  & $65,679$ & $3,420$ & $3,420$ & $6,601$ & $230$ & $365$\\
ICEWS14-irregular  & $29,102$ & $3,555$ & $3,607$ & $5,093$ & $210$ & $146$\\
\hline
\end{tabular}}

\end{table}


\section{Implementation details}
\label{sec:appendix c}

We implement all experiments with PyTorch \cite{DBLP:conf/nips/PaszkeGMLBCKLGA19} and use a single NVIDIA Tesla T4 for computation. We allow TARGCN to search for neighbors along the whole timeline. The hyperparameter searching strategies are reported in Table \ref{tab: hyperparameter search} and the hyperparameter settings producing the reported experimental results (in Table \ref{tab: tid link prediction results}) are presented in Table \ref{tab: Best hyperparameter setting}. We do 180 trials for each dataset and run the models for 20, 20 and 2 epochs on ICEWS14, ICEWS05-15 and GDELT, repectively. We choose the trial leading to the best MRR as the best hyperparameter setting. We use the official implementation of TComplEx, TNTComplEx \footnote{https://github.com/facebookresearch/tkbc}, TeMP \footnote{https://github.com/JiapengWu/TeMP} and T-GAP \footnote{https://github.com/sharkmir1/T-GAP}. We find that T-GAP has an extremely high memory demand. Training GDELT with T-GAP on a 16GB NVIDIA Tesla T4 causes out-of-memory error even when we set batch size to 1. This is due to its PGNN which constructs a huge temporal associative graph for every entity in training examples. 

The training time and the memory usage of TARGCN are reported in Table \ref{tab: computational budget}. The training time of TARGCN scales with the number of training quadruples in each dataset. Sampling temporal neighbors for every query subject requires relatively long computation time. This may cause timeout problems during the training process when TARGCN is used to train large-scale datasets (even much larger than GDELT). However, the memory usage of our model remains quite low, which enables training on smaller GPUs.

\begin{table}[ht!] 
    \caption{Hyperparameter searching strategy.}
    \label{tab: hyperparameter search}
    \begin{center}
      \resizebox{0.9\columnwidth}{!}{
    \begin{tabular}{lcccc} 
      \toprule 
     \multicolumn{1}{l}{Datasets} & \multicolumn{1}{c}{\textbf{ICEWS14}} & \multicolumn{1}{c}{\textbf{ICEWS05-15}} &\multicolumn{1}{c}{\textbf{GDELT}} \\
      \midrule 
      Hyperparameter &   &  &   \\
      \midrule 
      Embedding Size & \{150, 200, 300\}     &  \{150, 200, 300\} &  \{150, 200, 300\} \\
      \# Aggregation Step &  \{1, 2\}   &  \{1, 2\}&   \{1, 2\}\\
      Activation Function &  \{Tanh, ReLU\}&  \{Tanh, ReLU\}&  \{Tanh, ReLU\}\\
      Search Range & \{15, 100, 200, 300, 365\}& \{100, 500, 1000, 4017\}& \{100, 200, 366\}\\
      \# Temporal Neighbor & \{50, 100, 500\} & \{50, 100, 500\}& \{50, 100, 500\}\\
       
      \bottomrule 
    \end{tabular} }
    \end{center}
\end{table}

\begin{table}[ht!] 
    \caption{Best hyperparameter settings on each dataset.}
    \label{tab: Best hyperparameter setting}
    \begin{center}
      \resizebox{0.6\columnwidth}{!}{
    \begin{tabular}{lcccc} 
      \toprule 
     \multicolumn{1}{l}{Datasets} & \multicolumn{1}{c}{\textbf{ICEWS14}} & \multicolumn{1}{c}{\textbf{ICEWS05-15}} &\multicolumn{1}{c}{\textbf{GDELT}} \\
      \midrule 
      Hyperparameter &   &  &   \\
      \midrule 
      Embedding Size & 300     &  200&  200\\
      \# Aggregation Step &  1   &  1&   1\\
      Activation Function &  Tanh&  Tanh&  Tanh\\
      Search Range & 365& 4017& 366\\
      \# Temporal Neighbor & 100 & 100& 100\\
       
      \bottomrule 
    \end{tabular} }
    \end{center}
\end{table}

\begin{table}[ht!] 
    \caption{Computational budget of TARGCN on benchmark datasets.}
    \label{tab: computational budget}
    \begin{center}
      \resizebox{0.6\columnwidth}{!}{
    \begin{tabular}{lcccc} 
      \toprule 
     \multicolumn{1}{l}{Datasets} & \multicolumn{1}{c}{\textbf{ICEWS14}} & \multicolumn{1}{c}{\textbf{ICEWS05-15}} &\multicolumn{1}{c}{\textbf{GDELT}} \\
      \midrule 

      GPU Memory Usage (MB) & 1,375     &  1,385 &  1,261\\
      Train Time/ Epoch (s)  &  405   &  9,900 &   145,200\\
      \# Train Epochs & 100 & 100 & 10\\
       
      \bottomrule 
    \end{tabular} }
    \end{center}
\end{table}

\section{Validation results}
\label{sec:appendix d}

In Table \ref{tab: tid link prediction val results}, we report the experimental results of TARGCN on validation sets on all three benchmark datasets. The results are produced by the same trained models reported in Table \ref{tab: tid link prediction results}.

\begin{table}[htbp]
\caption{Temporal knowledge graph completion results on the validation sets of three benchmark datasets. Evaluation metrics are
    filtered MRR and Hits@1/3/10.}\label{tab: tid link prediction val results}
    \centering
    \resizebox{\textwidth}{!}{
    \large\begin{tabular}{@{}lcccccccccccc@{}}
\toprule
        \textbf{Datasets} & \multicolumn{4}{c}{\textbf{ICEWS14}} &  \multicolumn{4}{c}{\textbf{ICEWS05-15}} & \multicolumn{4}{c}{\textbf{GDELT}}\\
\cmidrule(lr){2-5} \cmidrule(lr){6-9} \cmidrule(lr){10-13}
        \textbf{Model} & MRR & Hits@1 & Hits@3 & Hits@10 & MRR & Hits@1 & Hits@3  & Hits@10 & MRR & Hits@1 & Hits@3  & Hits@10 \\
\midrule 
        TARGCN 
        & 0.647 & 0.591 & 0.679 & 0.748
        & 0.705  & 0.641 & 0.742 & 0.821
        & 0.510 & 0.418 & 0.552 & 0.685  
        \\
\bottomrule
    \end{tabular}
    }
    
\end{table}

\begin{table}[htbp]
\caption{Performance of TARGCN coupled with different KG score functions. Embsize means the embedding size of entity and relation representations.
    }\label{tab: kg score}
    \centering
    \resizebox{0.8\columnwidth}{!}{
    \large\begin{tabular}{@{}lcccccccccc@{}}
\toprule
        \textbf{Datasets} & \multicolumn{5}{c}{\textbf{ICEWS14}} &  \multicolumn{5}{c}{\textbf{ICEWS05-15}}\\
\cmidrule(lr){2-6} \cmidrule(lr){7-11}
        \textbf{Model} & MRR & Hits@1 & Hits@3 & Hits@10 & Embsize & MRR & Hits@1 & Hits@3 & Hits@10 & Embsize \\
\midrule 
        TARGCN + ComplEx  & 0.628 & 0.562 & 0.667 & 0.745 & 300 & 0.692 & 0.624 & 0.734 & 0.816 & 200
         \\
        TARGCN + BiQUE  & 0.629 & 0.561 & 0.666 & \textbf{0.753} & 320 & 0.701 & 0.634 & 0.739 & \textbf{0.824} & 200
         \\
        TARGCN + Distmult & \textbf{0.636} & \textbf{0.576} & \textbf{0.672} & 0.746 & 300 & \textbf{0.702} & \textbf{0.635} & \textbf{0.743} & 0.823 & 200
        \\

\bottomrule
    \end{tabular}
    }
    
\end{table}
\section{Comparison over different KG socre functions.}
\label{sec: appendix kg score}
We couple TARGCN with three different KG score functions, i.e., ComplEx \cite{DBLP:conf/icml/TrouillonWRGB16}, BiQUE \cite{guo-kok-2021-bique}, Distmult \cite{DBLP:journals/corr/YangYHGD14a}, and report their performances on ICEWS14 and ICEWS05-15 in Table \ref{tab: kg score}. All these KG score functions do not include additional parameters besides the entity and relation representations. TARGCN's number of parameters only scales with the embedding size, indicating that as long as the embedding size remains unchanged, there is no change in the number of parameters when TARGCN is couple with another KG score function, e.g., switching from Distmult to ComplEx. For TARGCN + ComplEx, we keep the embedding size of both entities and relations as same as the size in TARGCN + Distmult that generates the results in Table \ref{tab: tid link prediction results} (300 on ICEWS14 and 200 on ICEWS05-15, reported in Table \ref{tab: Best hyperparameter setting}). BiQUE requires that the embedding size is divisible by 8. Since 300 is not divisible by 8, we set the embedding size of TARGCN + BiQUE to 320 on ICEWS14, and to 200 on ICEWS05-15. 
From Table \ref{tab: kg score}, we observe that TARGCN + Distmult achieves the best performance on both datasets, even when TARGCN + BiQUE has more parameters on ICEWS14. To this end, we choose TARGCN + Distmult as our final model structure due to its high parameter efficiency. We also notice that TARGCN constantly shows strong performance when it is applied with different KG score functions. Though TARGCN + ComplEx performs the worst in Table \ref{tab: kg score}, it still outperforms previous TKGC methods on the benchmark datasets.

\section{Further details of generalization to unseen timestamps and irregular timestamped data}
\label{sec: appendix e}
We choose four strong baselines to compare with TARGCN, namely, TComplEx \cite{DBLP:conf/iclr/LacroixOU20}, TNTComplEx \cite{DBLP:conf/iclr/LacroixOU20}, TeMP-SA \cite{DBLP:conf/emnlp/WuCCH20}, and T-GAP \cite{DBLP:conf/kdd/JungJK21}. We choose TeMP-SA since it is reported with better results on ICEWS14 (newly created datasets are based on ICEWS14). We cannot perform unseen timestamps generalization with TeMP-SA since it requires the unavailable KG snapshot $\mathcal{G}(t_q)$ for every link prediction query $(s_q, r_q, ?, t_q)$.

\subsection{Unseen timestamps generalization}
We do not use the same unseen timestamps generalization datasets proposed in \cite{DBLP:conf/aaai/GoelKBP20} and \cite{DBLP:conf/kdd/JungJK21}, since they did not release their datasets. We follow \cite{DBLP:conf/aaai/GoelKBP20} and create ICEWS14-unseen by ourselves. We exclude every quadruple appearing on the 5th, 15th, and 25th day of each month in ICEWS14 to construct a new training set. We randomly split the excluded quadruples into validation and test sets. We make sure that every entity appearing in the validation and test sets is seen in the training set.

By comparing the results in Table \ref{tab: tid link prediction results} and Table \ref{tab: unseen}, we observe that the performance improvement of TARGCN becomes even much larger on ICEWS14-unseen than on the original dataset. TARGCN achieves a relative improvement of 21.94\% on MRR compared with T-GAP and TNTComplEx. More surprisingly, it also achieves a relative improvement of 43.09\% on Hits@1 compared with the strongest baseline T-GAP on unseen timestamps generalization. This proves the extremely strong robustness of our model to link inference at unseen timestamps. 

\subsection{Performance on irregular timestamped data}
We sample the KG snapshots from the original 
ICEWS14 dataset. The value of the time interval between every two neighboring snapshots can be randomly assigned either to 1, 2, 3, or 4. In this way, we create a dataset simulating that the TKG data is observed and collected at irregular-spaced timestamps. 

TARGCN enlarges the performance gap between itself and other baselines, compared with the results regarding TKGC on the original dataset reported in Table \ref{tab: tid link prediction results}. Besides, we observe that TeMP-SA and T-GAP outperform TNTComplEx on ICEWS14-irregular, while they perform worse on the original dataset. This is due to their time displacement temporal encoders which learn different temporal embeddings for different time intervals. For TARGCN, it employs a time difference temporal encoder that maps time-aware entity representations with the explicit value of time differences, thus being able to capture accurate temporal information provided by irregular timestamped data.


\section{Parameter efficiency analysis details}
\label{sec: appendix param eff}
\begin{table}[htbp]
\caption{Parameter efficiency comparison on ICEWS14. We adopt relative change to define the increase in parameter numbers and the drop in MRR.}\label{tab: efficiency ICEWS14}
    \centering
    \resizebox{0.5\columnwidth}{!}{
    \large\begin{tabular}{@{}lcccc@{}}
\toprule
        \textbf{Datasets} & \multicolumn{4}{c}{\textbf{ICEWS14}}\\
\cmidrule(lr){2-5}
        \textbf{Model} & Parameters & MRR & Parameter $\uparrow$ & MRR $\downarrow$  \\
\midrule 
        TARGCN  & 1229100 & 0.627 & - & - 
         \\
        T-GAP  & 1912350 & 0.610 & 55.59\% & 2.71\% 
         \\
        TeMP-SA & 1264640 & 0.607 & 2.89\% & 3.19\% 
         \\
        TeMP-GRU  & 1413408 & 0.601 &15.00\% & 4.15\% 
         \\

\bottomrule
    \end{tabular}
    }
    
\end{table}
Similar to Table \ref{tab: efficiency GDELT,ICEWS05-15}, Table \ref{tab: efficiency ICEWS14} summarizes the number of parameters as well as performance difference on ICEWS14. For TARGCN, the model producing results in Table \ref{tab: tid link prediction results} has more parameters than T-GAP and TeMP. Therefore, we decrease the embedding size of TARGCN to 150 so that its parameter number becomes the smallest among all models. We keep T-GAP and TeMP with their optimal parameter settings and compare them with TARGCN. From Table \ref{tab: tid link prediction results} and Table \ref{tab: efficiency ICEWS14}, we observe that even when we decrease the embedding size of TARGCN from 300 to 150, our model still performs well (MRR drops from 0.635 to 0.627), and it still outperforms T-GAP and TeMP on ICEWS14. TeMP variants show the worst performance, even when they have more parameters than TARGCN. T-GAP performs better than TeMP variants. However, it uses 55.59\% more parameters than TARGCN, while it is beaten with a 2.71\% performance drop.

All the points in Figure \ref{fig: efficiency 14} are based on the results in Table \ref{tab: efficiency ICEWS14 for fig}. Note that we control the number of parameters only by changing embedding size, without changing any other hyperparameters or model structures.

\begin{table}[htbp]
\caption{Experimental results as well as the number of parameters that lead to Figure \ref{fig: efficiency 14}. Underlined results are taken from Table \ref{tab: tid link prediction results}.}\label{tab: efficiency ICEWS14 for fig}
    \centering
    \resizebox{0.5\columnwidth}{!}{
    \large\begin{tabular}{@{}lccc@{}}
\toprule
        \textbf{Datasets} & \multicolumn{3}{c}{\textbf{ICEWS14}}\\
\cmidrule(lr){2-4}
        \textbf{Model} & Embedding Size & Parameters & MRR  \\
\midrule 
        TARGCN  & 100 & 799400 & 0.605 
         \\
         & 150 & 1229100 &0.627\\
         & 200 & 1678800 &0.629\\
         & \underline{300} & \underline{2638200} &\underline{0.636}\\
\midrule
        T-GAP  & 50 & 928675 & 0.582 
         \\
         & \underline{100} & \underline{1912350} &\underline{0.610}
         \\
         & 200 & 2951025 &0.596
         \\
\midrule
        TeMP-SA & 64 & 611840 & 0.595  
         \\
        & \underline{128} & \underline{1264640} &\underline{0.607}  
         \\
         & 256 & 2928640 &0.618
         \\

\bottomrule
    \end{tabular}
    }
    
\end{table}

\end{document}